\documentclass{article}

\PassOptionsToPackage{numbers, compress}{natbib}

\usepackage[preprint]{neurips_2024}

\usepackage[utf8]{inputenc} 
\usepackage[T1]{fontenc}    
\usepackage{hyperref}       
\usepackage{url}            
\usepackage{booktabs}       
\usepackage{amsfonts}       
\usepackage{nicefrac}       
\usepackage{microtype}      
\usepackage{xcolor}         

\usepackage{soul}

\usepackage{enumitem}
\usepackage{amsfonts, amsmath, amssymb} 
\usepackage[ruled,linesnumbered,vlined]{algorithm2e}
\definecolor{green}{RGB}{223, 255, 216} 
\usepackage{multirow}

\usepackage{graphicx}
\usepackage{wrapfig}
\usepackage[subrefformat=parens, labelformat=parens]{subfig}

\usepackage{caption}

\usepackage{enumitem}

\usepackage{tabularray}
\usepackage{booktabs}

\newcommand{{\model}}{\text{SEB}}

\title{Subword Embedding from Bytes Gains \\Privacy without Sacrificing Accuracy  and Complexity}

\author{%
  Mengjiao Zhang \\
  Department of Computer Science\\
  Stevens Institute of Technology\\
  \texttt{mzhang49@stevens.edu} \\
  \And
  Jia Xu \\
  Department of Computer Science\\
  Stevens Institute of Technology\\
  \texttt{jxu70@stevens.edu} \\
}

\begin{document}

\maketitle

\begin{abstract}
  While NLP models significantly impact our lives, there are rising concerns about privacy invasion. Although federated learning enhances privacy, attackers may recover private training data by exploiting model parameters and gradients. Therefore, protecting against such embedding attacks remains an open challenge. 
  To address this, we propose Subword Embedding from Bytes ({\model}) and encode subwords to byte sequences using deep neural networks, making input text recovery harder. Importantly, our method requires a smaller memory with $256$ bytes of vocabulary while keeping efficiency with the same input length. Thus, our solution outperforms conventional approaches by preserving privacy without sacrificing efficiency or accuracy. 
  Our experiments show {\model} can effectively protect against embedding-based attacks from recovering original sentences in federated learning. Meanwhile, we verify that {\model} obtains comparable and even better results over standard subword embedding methods in machine translation, sentiment analysis, and language modeling with even lower time and space complexity.

\end{abstract}

\section{Introduction}

Advances in Natural Language Processing (NLP), such as Large Language Models (LLMs), have made noticeable advancements in performance over the last decades, partially attributed to the large datasets available. Since most data are from users, their privacy concerns play an increasingly critical role, which is essential to building user trust, encouraging the responsible use of language data, protecting personal information, ensuring ethical use, and avoiding potential harm to individuals. 

\begin{figure*}
    \centering
        \subfloat[][]{
            \includegraphics[width=0.42\linewidth]{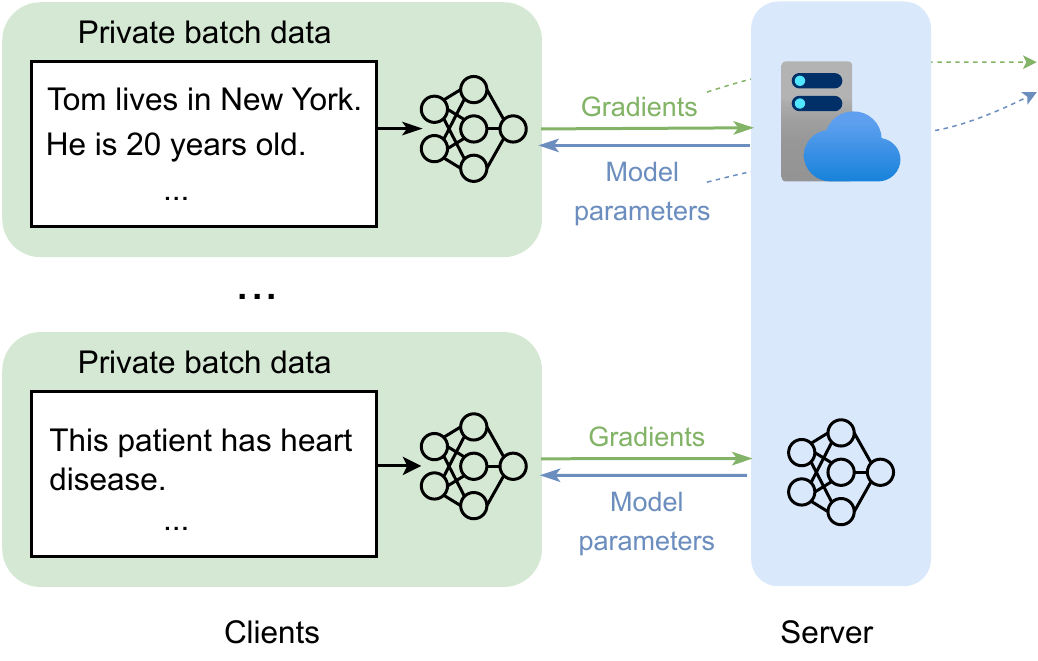}
            \label{fig: FL}
            }
            \hspace{-0.5cm}
        \subfloat[][]{
            \includegraphics[width=0.215\linewidth]{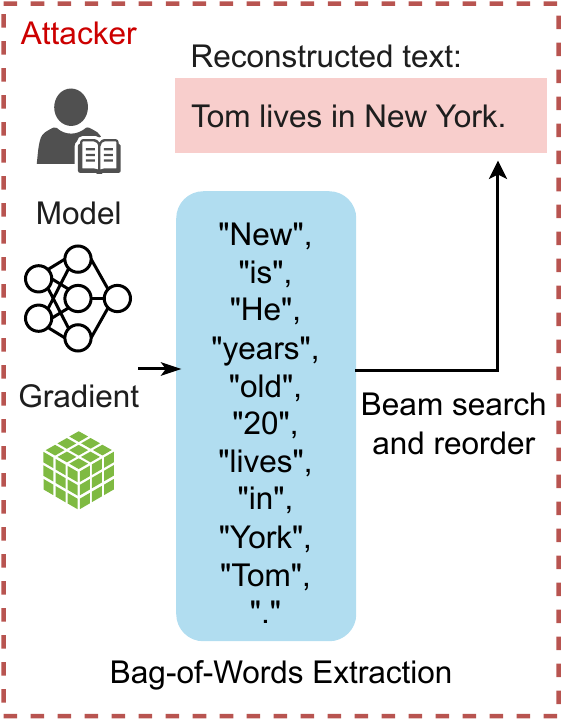}
            \label{fig: subword_attack}
            }
        \subfloat[][]{
            \includegraphics[width=0.33\linewidth]{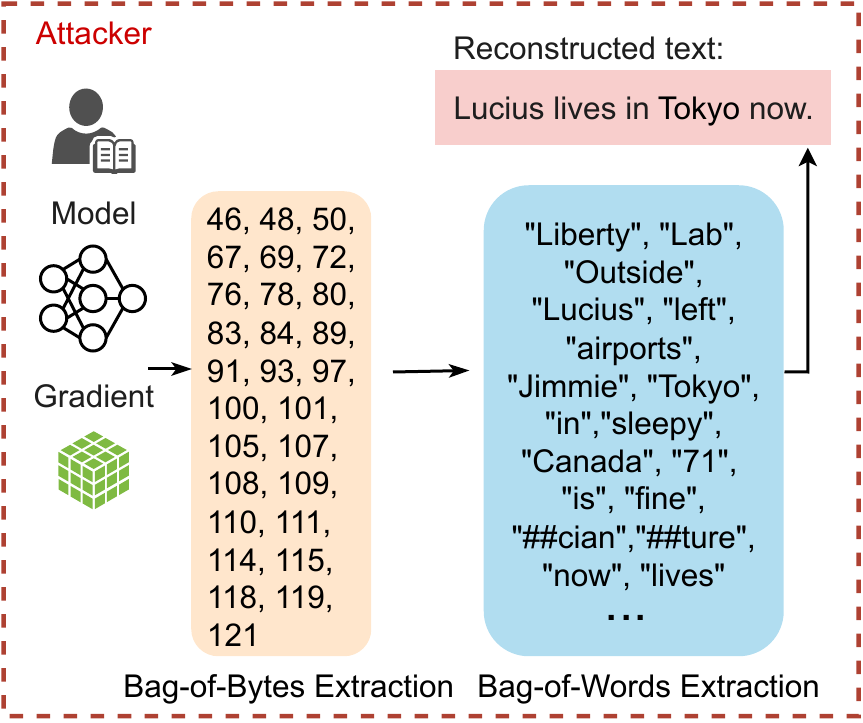}
            \label{fig: byte_attack}
            }
    \caption{An attack example of recovering text in FL. (a): An FL framework. (b) and (c): Recovering text using embedding gradients of subwords and bytes.}
    \label{fig: Fed_privacy}
\end{figure*}

Federated learning~(FL) enables training shared models across multiple clients without transferring the data to a central server to preserve user privacy. Although only the model updates are sent to the central server, adversaries can still use model updates to reconstruct the original data and leak sensitive information to compromise the user's privacy. 
\figureautorefname{~\subref*{fig: FL}} demonstrates an FL framework, and \figureautorefname{~\subref*{fig: subword_attack}} shows how embedding-based attacks work as in~\cite{gupta2022recovering}.
In the illustrated example, the attacker extracts all candidate words in a batch of data from the embedding gradients and can easily reconstruct the text with beam search and reordering since one can perform straightforward lookups when a vector is updated due to the one-to-one mapping between word/subword tokens 
and embedding vectors.  

Our intuitive idea is to apply the byte embedding method because the same bytes are repeatedly used for multiple subwords. We aim to design a one-to-many mapping between words/subwords and embedding vectors to increase the difficulty of the simple lookup so that retrieving input subwords with the updated byte embeddings is harder, which makes the byte embedding in NLP models a potential defense.
For example, in subword embedding, if the word ``good'' is updated, the attacker will only retrieve this word based on embedding updates. However, if we tokenize ``good'' into four bytes, such as ``50, 10, 128, 32", all subwords containing at least one of these bytes will be retrieved, resulting in a larger search space and more possibilities to recover the original sentence.
As shown in ~\figureautorefname{~\subref*{fig: byte_attack}}, although the attacker extracts a set of bytes, the number of candidate subwords is much greater than that of using subword embeddings. 

There are two major challenges to directly apply existing byte encodings~\cite{xue2022byt5, shaham2021neural, zhang-xu-2022-byte} to enhance privacy: 
First, smaller textual granularity cannot show the semantic meaning of each word, leading to a less interpretable and analyzable model. Second, byte-based models are more computationally expensive, as input sequences become much longer after byte tokenization.

To address these challenges in byte-based models, we propose to encode subwords with bytes and aggregate the byte embeddings to obtain a single subword embedding. 
The procedure consists of three steps:
(1) Construct a mapping between subwords and bytes.
(2) Convert the input text into a byte sequence.
(3) Retrieve the corresponding byte embeddings and aggregate them back into subword embeddings using a feed-forward network while maintaining the subword boundaries.
By adopting this approach, we can leverage the privacy protection provided by bytes with a small vocabulary size of $256$ while keeping the same input sequence length as the subword sequence. 

\textbf{Our main contributions are}:
\begin{itemize}[leftmargin=*]
    \item We introduce a novel text representation method {\model}, which achieves a vocabulary size of $256$ of the learned model without increasing the input sequence length.
    \item We verify that our {\model} can protect NLP models against data leaking attacks based on embedding gradients. To the best of our knowledge, our work is the first one to study privacy preservation with byte representations in FL.
    \item We demonstrate that {\model} improves privacy and, at the same time, achieves comparable or better accuracy with enhanced time and space efficiency without the privacy-performance/efficiency trade-off in conventional approaches. 
\end{itemize}

\section{Related Work}
\paragraph{Attacks and defenses in language model}

Some recent works consider the reconstruction as an optimization task \cite{zhu2019deep, deng2021tag, balunovic2022lamp}. The attacker updates its dummy inputs and labels to minimize the distance between the gradients of the victim uploaded and the gradients the attacker calculated based on its dummy inputs and labels.~\cite{gupta2022recovering} shows that the attackers can reconstruct a set of words with the embedding gradients, then apply beam search and reorder with a pretrained language model for input recovery. One defense described in~\cite{zhu2019deep, deng2021tag, balunovic2022lamp} is to encrypt the gradients or make them not directly inferable. However, encryption requires special setups and could be costly to implement. Moreover, it does not provide effective protection against server-side privacy leakage~\cite{aono2017privacy, huang2021evaluating, fang2021privacy}. Differential privacy is another defense strategy, but it can hurt model accuracy~\cite{zhu2019deep, wei2020framework,yin2021comprehensive, li2021large}. While \cite{zhang2021matrix} proposed a secure federated learning framework that can prevent privacy leakage based on gradient reconstruction, it does not effectively address the retrieval of a bag of words from the embedding matrix gradients, as proposed in \cite{gupta2022recovering}.

\paragraph{Subword-level and byte-level language models}
 Subword tokenization such as BPE~\cite{sennrich2015neural} has some limitations, despite the wide application. It cannot handle out-of-vocabulary subwords and requires language-specific tokenizers. Another challenge is the high space complexity of the embedding matrix when the vocabulary size is huge. Byte tokenization is a solution to address these issues~\cite{shaham2020neural, zhang-xu-2022-byte, xue2022byt5}. UTF-8 can encode almost all languages. Therefore, there will be no out-of-vocabulary words and the language-specific tokenizer is unnecessary. In addition, as the total number of bytes in UTF-8 is 256, the embedding matrix for byte vocabulary is much smaller than most subword vocabularies, reducing the number of parameters in the embedding layer and saving memory space.

\paragraph{Subword-level model with character- or byte-level fusion}
The character/byte-based models often result in longer input sequences and higher time complexity compared to the subword-based model. To make the model efficient, recent works have explored character/byte-level fusion. For example, \cite{tay2021charformer} proposes CHARFORMER, using a soft gradient-based subword tokenization module to obtain ``subword tokens''. It generates and scores multiple subword blocks, aggregates them to obtain subword representation, and then performs downsampling to reduce the sequence length. Although CHARFORMER is faster than vanilla byte/character-based models, it does not maintain subword boundaries, limiting the model's interpretability. \cite{sreedhar2022local} proposes Local Bytes Fusion (LOBEF) to aggregate local semantic information and maintain the word boundary. However, it does not reduce the sequence length, making training and inference time-consuming.

\section{Preliminaries}

\subsection{Federated Learning}
In federated learning (FL), multiple clients jointly train a model using their private data. Assume we have $N$ clients, $\mathcal{C} = \{c_1, c_2, \dots, c_N\}$, and a server $s$, in an FL system. The jointly trained model is $f$ with parameters $\theta$. The clients' private data are $\mathcal{D}_1, \mathcal{D}_2, \dots, \mathcal{D}_N$ and the objective function is $\mathcal{L}$. For easier illustration, we assume all clients participate in each communication and use FedSGD~\citep{mcmahan2017communication} to update the model parameters. In each communication round $t$, server $s$ first sends the model parameters $\theta ^ t$ to all clients. Then each client $c_i$ compute $\nabla_{\theta^t} \mathcal{L}_{\theta^t}(\mathcal{B}_i)$, the gradients of current model $f_{\theta ^ t}$, based on a randomly sampled data batch $\mathcal{B}_i \subset \mathcal{D}_i$. After local computation, the clients send the gradients $\Delta_1 ^{t}, \Delta_2 ^{t}, \dots, \Delta_N ^{t}$ to server and server $s$ aggregate all the gradients and update the model:
\begin{equation}
\label{eq:FedAvg}
    \theta^{t+1} = \theta^{t} - \eta \sum_{i=1} ^{N} \nabla_{\theta^t} \mathcal{L}_{\theta^t}(\mathcal{B}_i).
\end{equation}
Here, Equation~(\ref{eq:FedAvg}) is the gradient descent, and $\eta$ is the learning rate. 

\subsection{Threat Model}

\paragraph{Adversary's capabilities and objective}
We follow the attack settings in~\citep{gupta2022recovering}. The optimized model is a language model $\mathcal{L}$, parameterized by $\theta$. This scenario makes the attacker white box access to the gradients $\nabla_{\theta^t} \mathcal{L}_{\theta^t}(\mathcal{B}_i)$ sent by the victim client $c_i$. $\theta^t$ is the model parameter that the server sends to the clients at any communication round $t$. From parameters $\theta^t$ and gradients $\nabla_{\theta^t} \mathcal{L}_{\theta^t}(\mathcal{B}_i)$, the attacker can get the information of the vocabulary $\mathcal{V}$ and the embedding matrix $\mathbf{W}$ to retrieve which tokens are updated. The goal of the attacker is to recover at least one sentence from $\mathcal{B}_i$, based on $\nabla_{\theta^t} \mathcal{L}_{\theta^t}(\mathcal{B}_i)$ and $\theta^{t}$.




\paragraph{Attack model}

This paper does not address the gradient leakage attack which aims to obtain private data by minimizing the difference between gradients derived from a dummy input and the actual gradients of the victim's data, because several methods have been proposed to mitigate this particular attack~\cite{zhu2019deep, deng2021tag, wei2020framework}. Instead, we focus on a specific attack model, FILM, introduced in \cite{gupta2022recovering}, for which effective defenses have yet to be explored. In this model, the attacker attempts to reconstruct sentences from the victim's training batches as follows: (1) extracting candidate tokens from the gradients, (2) applying beam search with a pre-trained Language Model, such as GPT-2, to reconstruct the input sentence, and (3) reordering the subword tokens to achieve the best reconstruction.

\section{Proposed Method}
\label{sec: proposed-method}
\begin{figure*}
    \centering
        \subfloat[][]{
            \includegraphics[width=0.32\linewidth]{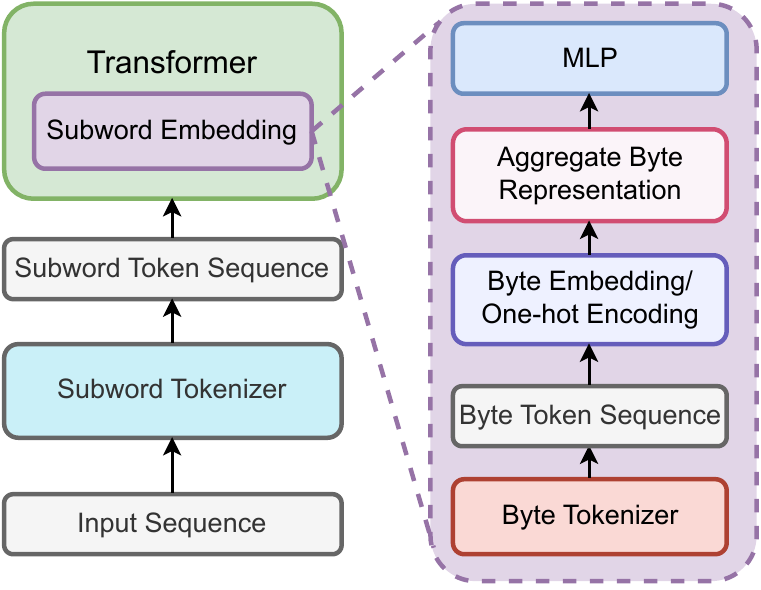}
            \label{fig: model_overview}
            }
            \hfill
        \subfloat[][]{
            \includegraphics[width=0.63\linewidth]{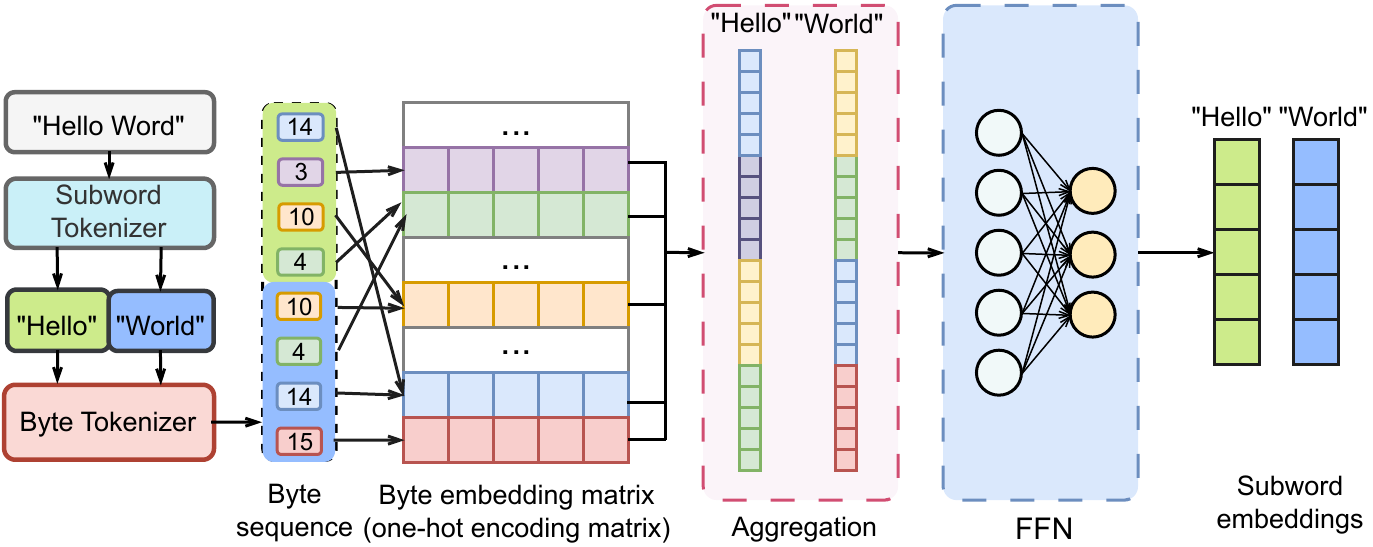}
            \label{fig: byte2word}
            }
    \caption{(a): An overview of the transformer model with {\model}. (b): An example of calculating subword embeddings with byte embedding.}
    \label{fig:model_arch}
\end{figure*}

We propose Subword Embedding from Bytes ({\model}) shown in \figureautorefname{~\ref{fig:model_arch}}, including byte sequence for input text, byte embeddings, aggregation of byte embeddings, and a feed-forward network to output the subword embedding.

We aim to develop subword encoding using a smaller byte embedding matrix to save space and protect against attacks based on the embedding gradients while preserving subword boundaries to maintain the model's time efficiency. This raises two main challenges: 1) how to convert subwords into a byte sequence? and 2) how to obtain subword embeddings using byte representations?

\subsection{Subword to Byte Sequence Mapping}
\label{sec: byte_tokenization}

UTF-8 encoding results in different sequence lengths for subwords. In real practice, all byte sequences need to be padded to the same length, making the byte sequence of the subword even longer. 
Instead of using the existing byte encoding system, we define our subword to byte sequence mapping $\mathcal{M}: \mathcal{V}_w \rightarrow (\mathcal{V}_b) ^ {n}$. $\mathcal{V}_w $ and $\mathcal{V}_b$ are subword and byte vocabularies with size of $V_w$ and $V_b$, respectively. $(\mathcal{V}_b) ^ {n}$ is a sequence of $n$ bytes in $\mathcal{V}_b$. Here the byte vocabulary size $V_b$ and the number of bytes $n$ to represent a subword are hyperparameters. In this way, every subword is represented with the same length, getting rid of the longer byte sequence with padding.

To construct the mapping, for every subword $w_i \in \mathcal{V}_w$, we randomly sample $n$ bytes with replacement from  $\mathcal{V}_b$ to obtain the byte sequence $(b_{i1}, b_{i2}, \dots, b_{in})$. If the byte sequence already exists in $\mathcal{M}$, we repeat the sampling until a unique byte sequence is obtained. For example, we set $V_b=64$ and $n = 4$. A subword ``Hello'' can be represented with $(14, 3, 10, 4)$, shown in \figureautorefname{~\subref*{fig: byte2word}}.

We analyze the probability $p$ that two subwords are mapped to the same byte sequence. With the byte vocabulary size $V_b$ and the number of bytes per subword $n$, the probability $p = 1/(V_b)^n$. For example, if $V_b = 16$ and $n=4$ then $p = 1.5 \times 10 ^{-5}$. For $V_b = 128$ and $n=8$ in our experiment, $p = 1.39 \times 10^{-17}$, which means there is almost no possibility to map two words into the same subword sequence. Therefore, {\model} is highly expressive for representing subwords.

\subsection{Subword Embedding from Bytes}
\label{sec:aggregation}

Different from the byte-based models which have higher time complexity due to longer input sequences, our method tokenizes the text into a sequence of bytes while preserving the subword boundary. We first tokenize the original text into subwords using a common subword tokenization method such as BPE. Then, we token each subword into a byte sequence with the mapping we designed above and then aggregate byte representations back to subword embeddings. The two detailed algorithms are in Appendix, Algorithms \ref{alg:byte_tokenizer} and~\ref{alg:aggregation}.

Let the byte embedding matrix be $\mathbf{B} \in \mathbb{R} ^{V_b \times d}$, where $d$ is the embedding size. Given a text $S$, we tokenize $S$ into a subword sequence $(w_1, w_2, \dots, w_m)$, then further use the mapping $\mathcal{M}$ defined above to tokenize this sequence into a byte sequence $(b_{11}, \dots, b_{1n}, \dots, b_{m1}, \dots, b_{mn})$ with $mn$ bytes. We retrieve the byte embeddings $\mathbf{E} \in \mathbb{R} ^{mn \times d}$ for these bytes from $\mathbf{B}$. 

To get a subword embedding, adding the byte representations for every $n$ bytes in $\mathbf{E}$ is a simple way. However, this approach does not consider the position of each byte within the subword. Considering that incorporating positional information can improve model performance for subword tokens, we induce positional information for byte sequences of subwords by concatenation. This enables the model to capture the position of each byte within the subword and obtain a more accurate and informative representation of the subword. Given the retrieved byte embeddings $\mathbf{E} \in \mathbb{R} ^{mn \times d}$, we reshape $\mathbf{E}$ to $\Tilde{\mathbf{E}} \in \mathbb{R} ^{m \times nd}$ in a row-major order, which is equivalent to concatenation. Then, an FFN is applied to project $\Tilde{\mathbf{E}}$ into the dimension $d'$ of the original subword embedding for language models: ${\mathbf{E}'} = \texttt{FFN}(\Tilde{\mathbf{E}}) \in \mathbb{R} ^{m \times d'}$. Note that, the byte embedding matrix $\mathbf{B}$ can be either a real-valued or one-hot embedding matrix because the vocabulary size is small for bytes. We compare the performances for both embedding methods in experiments.

\subsection{Complexity Analysis and Frequency Analysis}

\begin{wraptable}[9]{r}{0.45\textwidth}
\vspace{-1.2\baselineskip}
\small
\centering
\caption{Complexity for conventional subword embeddings, byte embedding, and our proposed {\model}. }
\begin{tabular}{lll} 
\toprule
Embedding  & Memory & Time \\
\midrule
Subword & $\mathcal{O}(V_wd)$ & $\mathcal{O}(m^2d)$ \\
Byte & $\mathcal{O}(V_bd)$ & $\mathcal{O}(c^2m^2d)$ \\
{\model} (Ours) & $\mathcal{O}((nd + V_b)d)$ & $\mathcal{O}(m^2d)$ \\
\bottomrule
\end{tabular}
\label{tab:complexity}
\end{wraptable}

\begin{figure*}
     \centering
     \subfloat[Batch size = 1]{
        \includegraphics[width=0.3\linewidth]{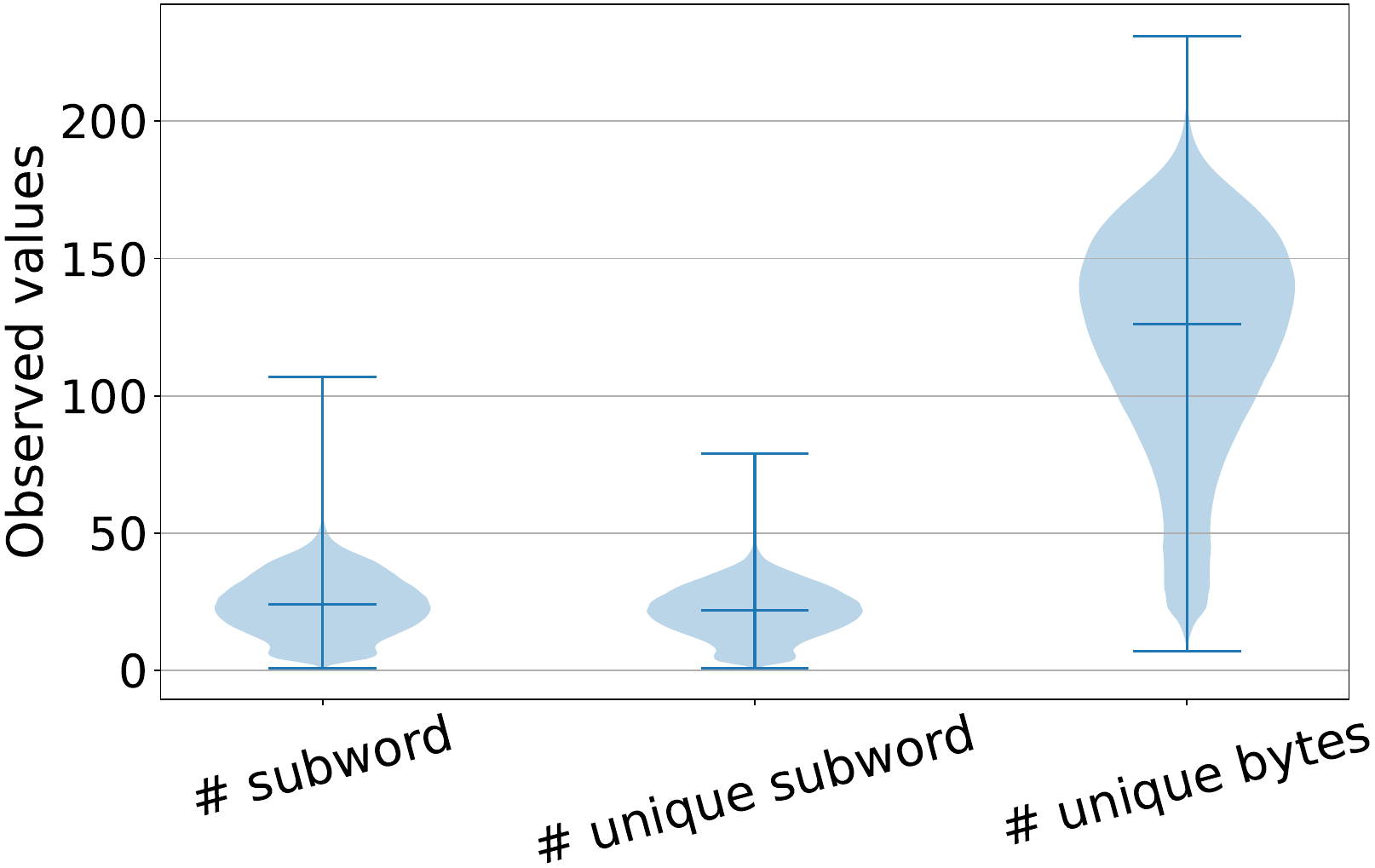}
        \label{fig: bs1}}
    \subfloat[Batch size = 4]{
        \includegraphics[width=0.3\linewidth]{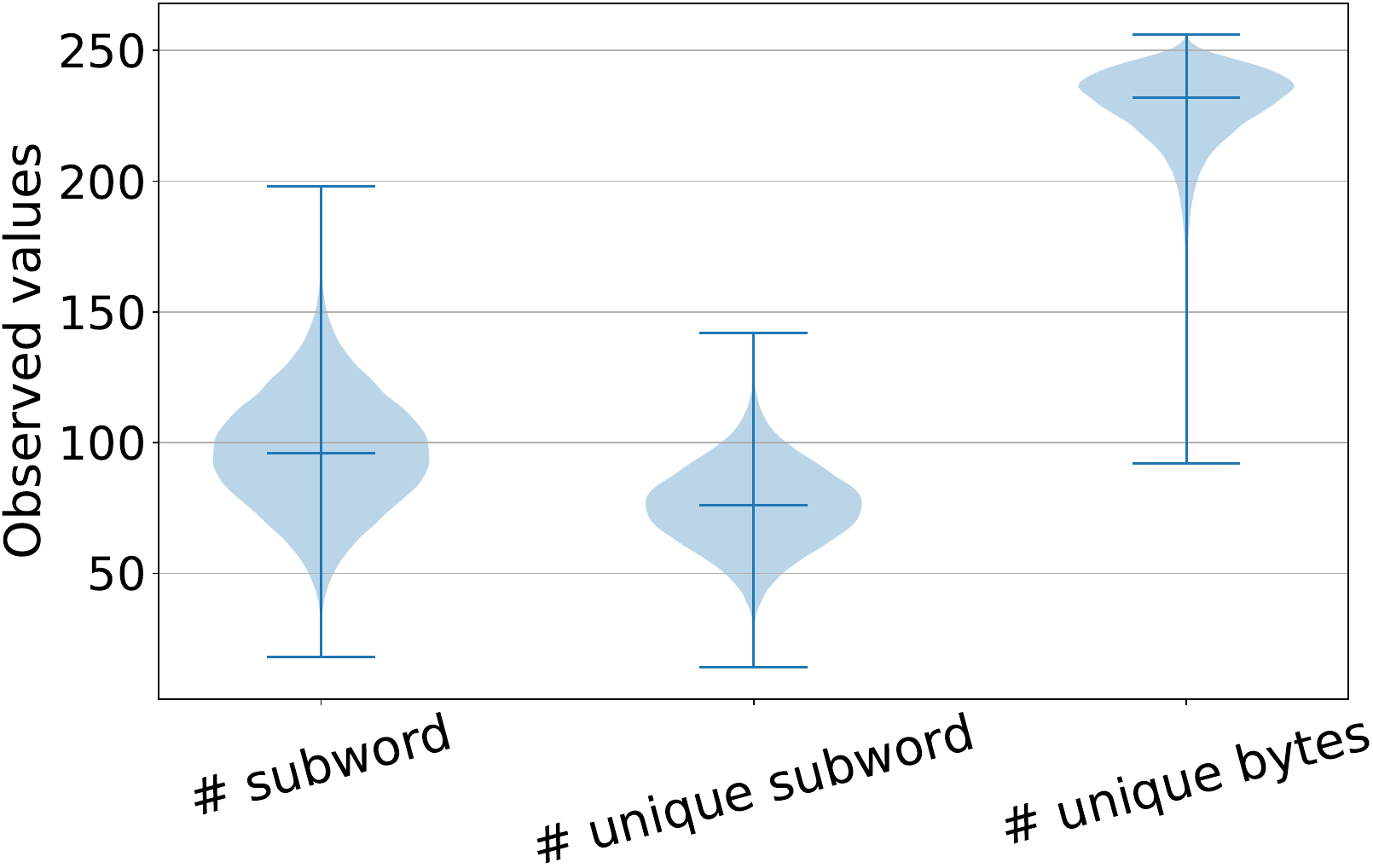}
        \label{fig: bs4}}
    \subfloat[Batch size = 16]{
        \includegraphics[width=0.3\linewidth]{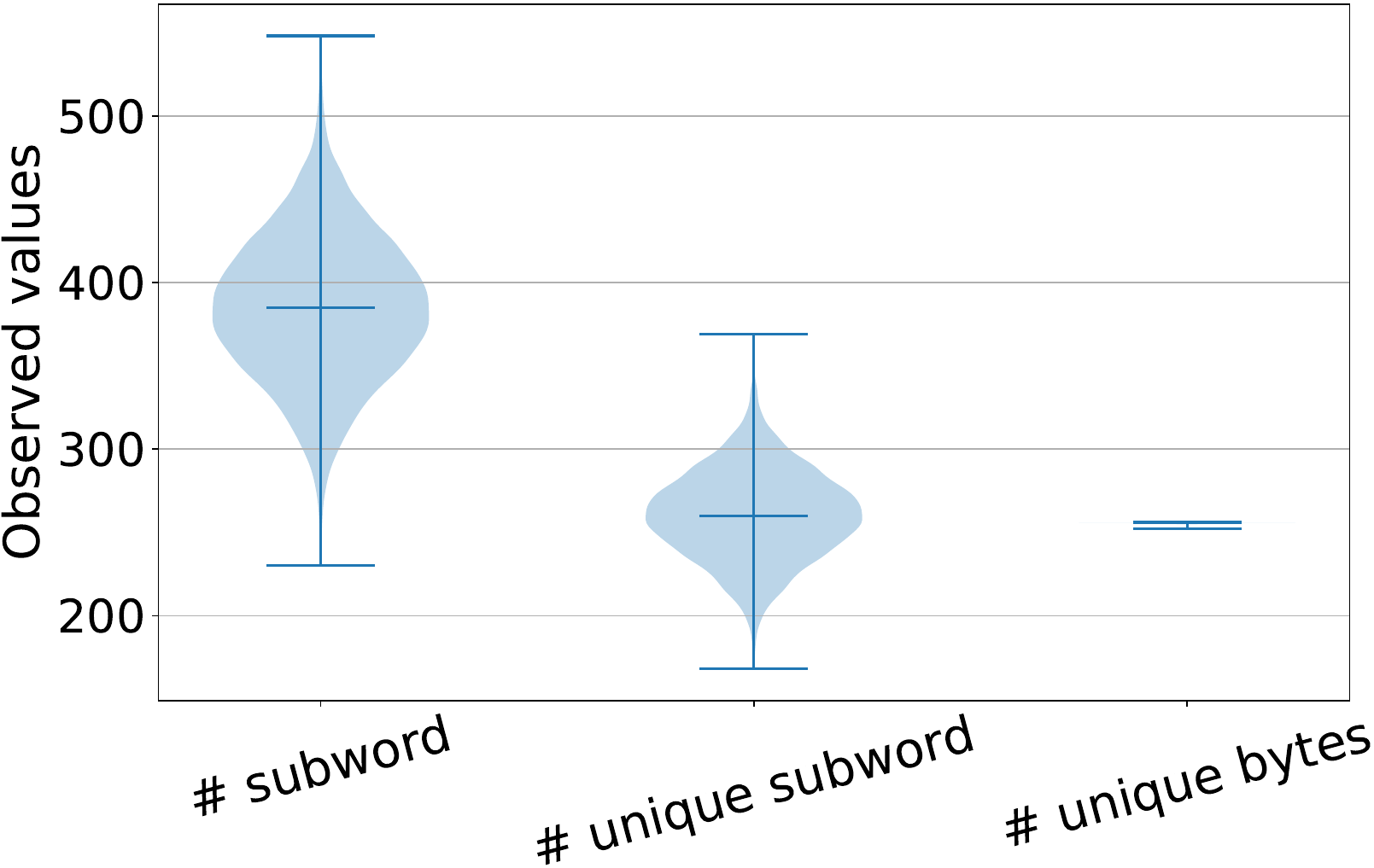}
        \label{fig: bs16}}
    \caption{The distribution of subword number, unique subword number, and unique byte number in a batch when batch size is 1, 4, 16. The vocabulary sizes of subwords and bytes are 50K and 256. }
    \label{fig: distribution}
\end{figure*}

To demonstrate the efficiency of the proposed {\model}, we summarize the space and time complexity of each embedding method in Table \ref{tab:complexity}. Here, the column ``Memory'' represents the memory usage for each embedding, and the column ``Time" shows the time complexity in Transformer attention. For simplicity, we let $d' = d$ and use one linear layer as FFN in {\model} which contains $nd^2$ parameters.

In terms of space complexity, subword embeddings typically have an exponentially large vocabulary size $V_w$, exceeding $10^4$, while the dictionary size of byte embeddings is no more than 256. For the proposed {\model}, the number of parameters in embedding is $\mathcal{O}(nd^2 + V_bd) = \mathcal{O}((nd + V_b)d)$, including the FFN and byte embedding matrix. In practice, $nd + V_b \ll V_w$.
As a result, both byte embeddings and our proposed {\model} significantly reduce the memory cost required for embeddings. In~\ref{app:space}, we show the analysis for space complexity in our experiments.
Regarding time complexity, we analyze the attention in the widely used Transformer~\cite{vaswani2017attention}. Given the sequence length $m$, byte embedding is more time-consuming since the input length is $c$ times longer than subword embedding. Here $c$ is the average ratio between the lengths of byte and subword sequences. Based on the statistics~\cite{shaham2020neural}, $c$ is usually around 5. However, our proposed {\model} maintains the same time efficiency as conventional subword embeddings because we preserve the subword sequence along with its boundaries. 

One may consider the frequency analysis in cryptanalysis to get the original text if the attacker also has information about the tokens used for the text and the frequency of each byte. In Appendix~\ref{app:freq_analysis}, we discuss the frequency analysis does not applicable to our proposed defense.

\section{Experiment}
\label{experiment}
We conduct experiments to demonstrate the advantages of {\model} in reducing space complexity, maintaining time efficiency, and preserving privacy for NLP models in federated learning. In all experiments, we set $V_{b} = 256$ and $n=8$, which is sufficient to prevent encoding two subwords into the same byte sequences. We use a 2-layer FFN in the proposed {\model}.


\subsection{Experiments on Privacy Protection}
\label{sec:expe_defense}

\paragraph{Dataset, attack task, and evaluation metrics}
We followed the settings in the FILM attack ~\citep{gupta2022recovering}. The dataset is WikiText-103~\cite{merity2016pointer}. For the attack task, we use GPT-2 base~\cite{radford2019language} with 117M parameters to recover the input batches. The ROUGE-1/2/L F-Scores~\cite{lin-2004-rouge} are used to evaluate the similarity between the recovered and original text.

\paragraph{Quantitative analysis of defense}

\begin{wrapfigure}[14]{r}{0.33\textwidth}
\vspace{-1\baselineskip}
    \centering
    \includegraphics[width=\linewidth]{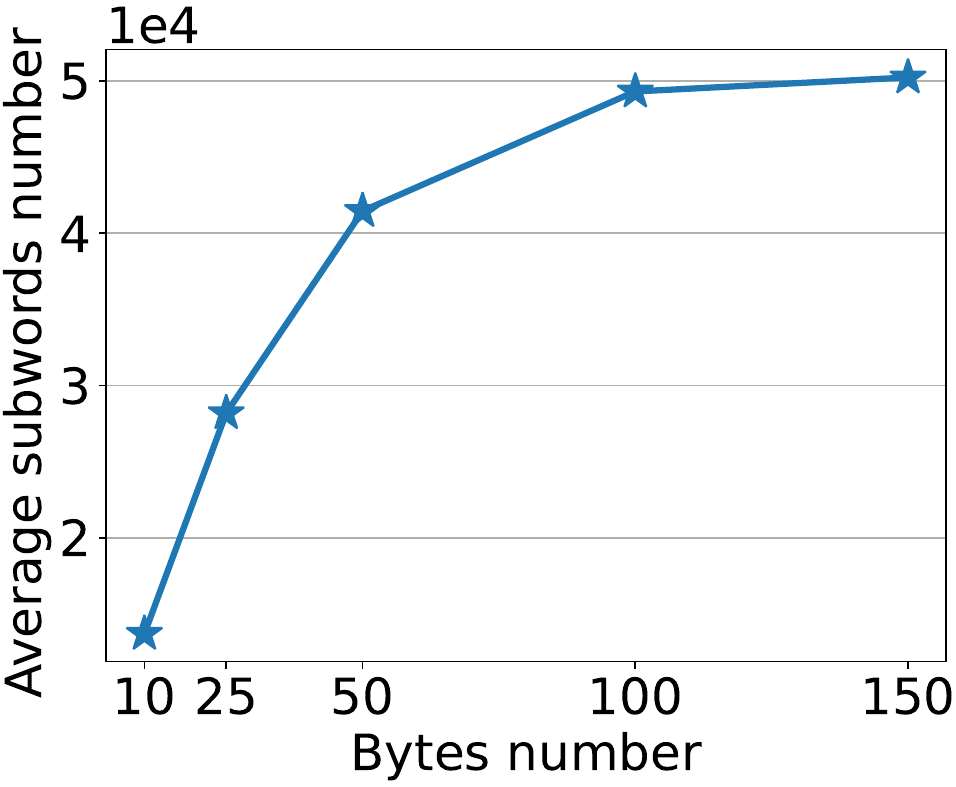}
    \caption{The average coverage of subwords given a random set of bytes with GPT-2 tokenizer.}
    \label{fig:byte2word_coverage}
\end{wrapfigure}

We first show that it is difficult to retrieve a bag of candidate subwords in {\model} with \figureautorefname{~\ref{fig: distribution}} and \ref{fig:byte2word_coverage}. In \figureautorefname{~\ref{fig: distribution}}, we present the distributions of the subword number, unique subword number, and unique byte number in a client's batch of data. We observe that even a single sample contains over 120 unique bytes on average, while only having approximately 25 unique subwords. In  \figureautorefname{~\ref{fig:byte2word_coverage}}, we present the average coverage of subwords for a subset of bytes. Based on \figureautorefname{~\ref{fig:byte2word_coverage}}, 120 bytes cover about 50K subwords. It means recovery is a random generation using almost the entire vocabulary.

\begin{figure}
     \centering
     \begin{minipage}[c]{0.84\linewidth}
        \subfloat[ROUGE-1]{
        \centering
        \includegraphics[height=0.175\linewidth]{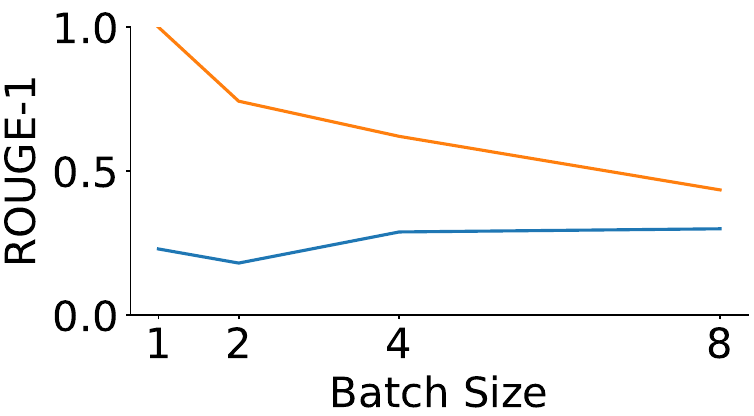}
        \label{fig: rouge1}}
    \subfloat[ROUGE-2]{
        \centering
        \includegraphics[height=0.175\linewidth]{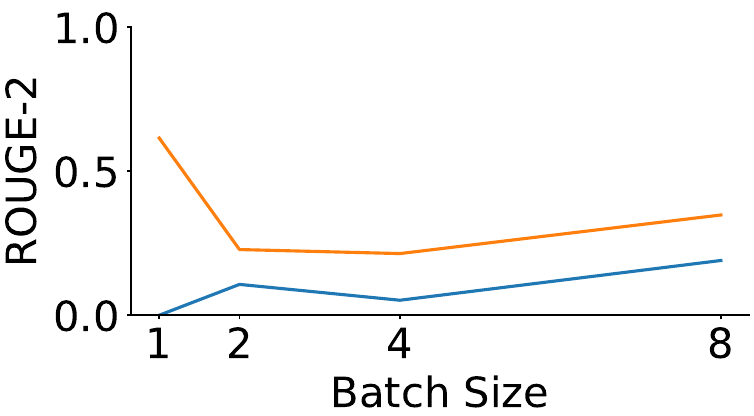}
        \label{fig: rouge2}} 
    \subfloat[ROUGE-L]{
        \centering
        \includegraphics[height=0.175\linewidth]{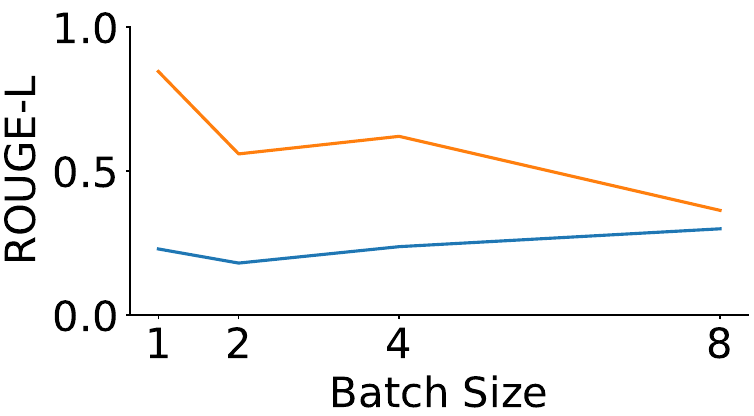}
        \label{fig: rouge3}} 
    \end{minipage}
    \begin{minipage}[c]{0.12\linewidth}
        \subfloat{
            \includegraphics[width=0.9\linewidth]{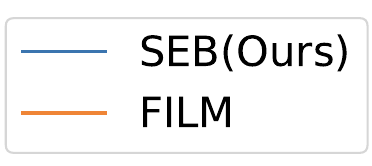}}
    \end{minipage}
    \caption{Recovery performance for batch size 1, 2, 4, 8 on WikiText-103.}
    \label{fig:recovery_rouge_socre}
\end{figure}

Additionally, \figureautorefname{~\ref{fig:recovery_rouge_socre}} shows the FILM attack performances using various batches on WikiText-103, with subword embedding and {\model}. As the candidate subwords are almost the whole vocabulary, beam search takes huge memory which is not executable on our device. To show the defense performance, we lose the constraints and randomly sample 7,000 subwords, combined with the subwords in the original text. We randomly select 5 tested batches for each batch size and take the average ROUGE F-Scores. When batch size is 1, ROUGE-1/2/L scores are close to 1 for attacks with subword embedding, indicating a nearly perfect recovery. However, these scores are quite low when using {\model}, showing the effectiveness of {\model} to defend the attacks based on embedding gradients. 


\paragraph{Qualitative analysis of defense} To intuitively show the difference between the recovered sentences of FILM using subword embedding and the proposed {\model}, we select the best-recovered sentences of these two methods based on the ROUGE-L F-score and list the results in 
 \tableautorefname{~\ref{tab: recovery}}. In the recovered sentence with the subword embedding, all words are successfully retrieved and have a very close order to the original sentence. However, with {\model}, only a few words are retrieved, and many of them are stop words. The results show that {\model} can prevent the attacker from recovering private information in the original sentence even though the batch only contains one sentence. 

 We also compare our method with the defense method of gradient pruning in the FILM attack \citep{gupta2022recovering} for batch size $= 8, 16, 32$. The results are shown in Appendix \ref{app:gradient_defense}.


\sethlcolor{green}
\begin{table*}
\small
    \centering
    \caption{The best recovered sentences by FILM using subword embedding and \model with batch size 1. Text in \hl{green} are successfully recovered phrases and words. \\}
    \begin{tabular}{lp{0.36\textwidth}p{0.45\textwidth}}
        \toprule
        \textbf{} & \textbf{Original Sentence} & \textbf{Best Recovered Sentence} \\ 
        \midrule
        Subword & The historic rainfall caused several dams to fill throughout northeast Mexico. & \hl{The} \hl{rainfall caused several} \hl{historic} \hl{dams} \hl{to fill throughout northeast Mexico}. \\
        \midrule
        {\model} & Pujols is a highly regarded hitter who has shown a "combination of contact hitting ability, patience, and raw power" & He \hl{is a} professional who \hl{has} \hl{a} very high degree of \hl{ability}, and always takes great advice, without ever assuming \hl{power}" ("Pivotal Decision Making With Your Head". Retrieved 12 Dec 2007 16 Mar) \\
        \bottomrule
    \end{tabular}
    \label{tab: recovery}
\end{table*}

\subsection{Experiment on Performance}
To provide a more comprehensive assessment of our proposed technique's applicability and performance of federated learning and across different NLP tasks, we conduct experiments on machine translation, sentiment analysis, and Language Modeling~(LM) tasks. The environmental settings are described in Appendix~\ref{app:env}.

\subsubsection{Translation}

\paragraph{Dataset and evaluation metrics}
In the translation task, we consider two datasets, the medium-size dataset IWSLT14~\citep{cettolo2014report} and the large-scale dataset WMT14~\cite{bojar-etal-2014-findings}. 
We follow the settings as prior work~\citep{shaham2020neural, zhang-xu-2022-byte} and translate German (de) to English (en) in IWSLT14~\citep{cettolo2014report}. The translation of WMT is English (en) to German (de) and the preprocessing is the same as Fairseq~\citep{ott2019fairseq}. 
We use SacreBLEU, case-sensitive, with the 13a tokenizer~\cite{post-2018-call} as the evaluation metric. A detailed description of preprocessing, model architecture, and hyperparameter settings can be found in Appendix~\ref{App:pre}.

\paragraph{Main results}
\begin{wraptable}[14]{r}{0.52\textwidth}
\vspace{-1\baselineskip}
\small
    \small
    \centering
    \caption{BLEU score of IWSLT14 and WMT14. {\model}$_{ar}$ and {\model}$_{cr}$: {\model} with added and concatenated real-valued embeddings, respectively. {\model}$_{co}$: {\model} with concatenated one-hot byte embeddings.}
    \begin{tabular}{@{}llll@{}}
    \toprule
    Datasets & Embeddings & \# Params & BLEU \\ \midrule
    \multirow{4}{*}{IWSLT14} & Subword & 5.2M & 34.54 $\pm$ 0.10 \\
     & {\model}$_{ar}$ & 4.3M & 34.64 $\pm$ 0.15 \\
     & {\model}$_{cr}$ & 9.6M & 35.32 $\pm$ 0.15 \\
     & {\model}$_{co}$ & 5.2M & \textbf{35.44 $\pm$ 0.10} \\ \midrule
    \multirow{2}{*}{WMT14} & Subword & 22.3M & 26.0 \\
     & {\model}$_{co}$ & 6.3M & 26.0 \\ \bottomrule
    \end{tabular}
    \label{tab:translation}
\end{wraptable}
For IWSLT14, we run 5 trials and report the average performance with the standard deviation. We show the translation results of Transformer with subword embedding and {\model} in Table \ref{tab:translation}. The hidden dimension of the two-layer FFN is 2048 for IWSLT because we try to keep the total parameters of {\model}$_{co}$ the same as the original Transformer. For WMT, the hidden dimension of FFN is 4096. Here, we test three variants of {\model} when aggregating the byte embedding back to subword embedding: added real-valued embedding ({\model}$_{ar}$), concatenated real-valued embedding ({\model}$_{cr}$), and concatenated one-hot embedding ({\model}$_{co}$). In this experiment, the dimensions of real-valued and one-hot vectors are 512 and 256. 
Table \ref{tab:translation} shows that {\model}$_{cr}$ and {\model}$_{co}$ can achieve better performances than subword embedding. 
Concatenating the one-hot vectors yields better results even with fewer model parameters than concatenating byte embedding. Therefore, we can conclude that {\model} is a better alternative to using large subword embeddings. Additionally, based on the comparison between {\model}$_{ar}$ and {\model}$_{cr}$, we find that concatenation is better than the simple adding of byte embeddings. This is expected as Section~\ref{sec:aggregation} because adding does not consider the positional information of bytes. The result of WMT14 shows the same performance as the subword-based model but with a smaller size of embedding parameters. It is important to emphasize that while privacy is improved, our model achieves the same or better accuracy than the baselines.

\begin{wraptable}{r}{0.45\textwidth}
\small
\vspace{-1\baselineskip}
    \centering
    \caption{Number of embedding parameters.}
    \begin{tabular}{cccc} 
    \toprule
    \multirow{2}{*}{\begin{tabular}[c]{@{}c@{}}Byte Tokens\\per Subword~($n$)\end{tabular}} & \multicolumn{3}{c}{Byte Vocabulary Size~($V_b$)}  \\ 
    \cmidrule{2-4}
                                                                                      & 64    & 128   & 256                       \\ 
    \midrule
    4                                                                                 & 0.79M & 1.05M & 1.57M                     \\
    8                                                                                 & 1.05M & 1.57M & 2.62M                     \\
    16                                                                                & 1.57M & 2.62M & 4.72M                     \\
    \bottomrule
    \end{tabular}
    \label{tab:byte_dict_para}
\end{wraptable}

\paragraph{Sensitivity analysis on $V_b$ and $n$}

\begin{figure*}
     \centering
     \subfloat{
        \includegraphics[width=0.46\linewidth]{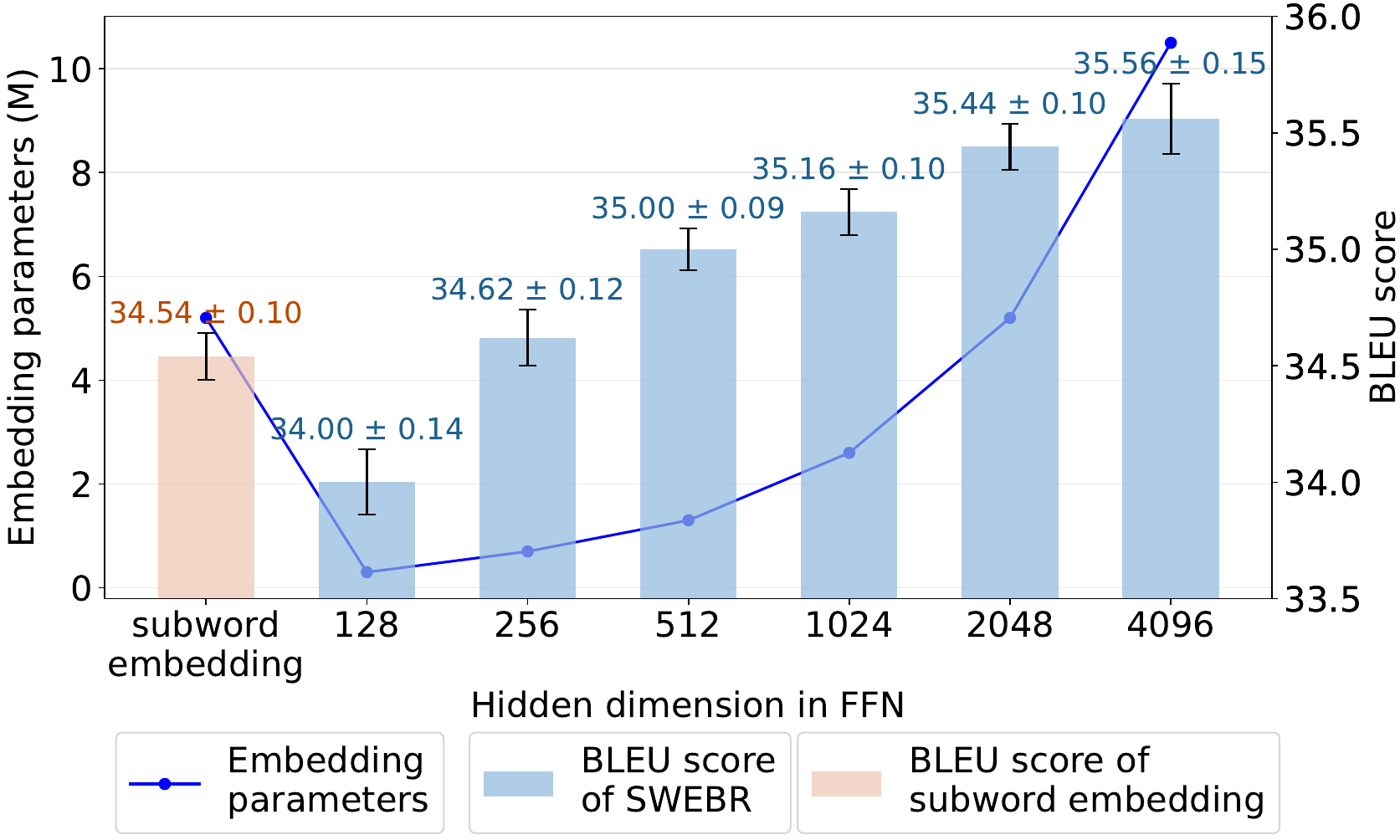}
        \label{fig: BLEU_hidden}}
    \quad\quad
    \subfloat{
        \includegraphics[width=0.38\linewidth]{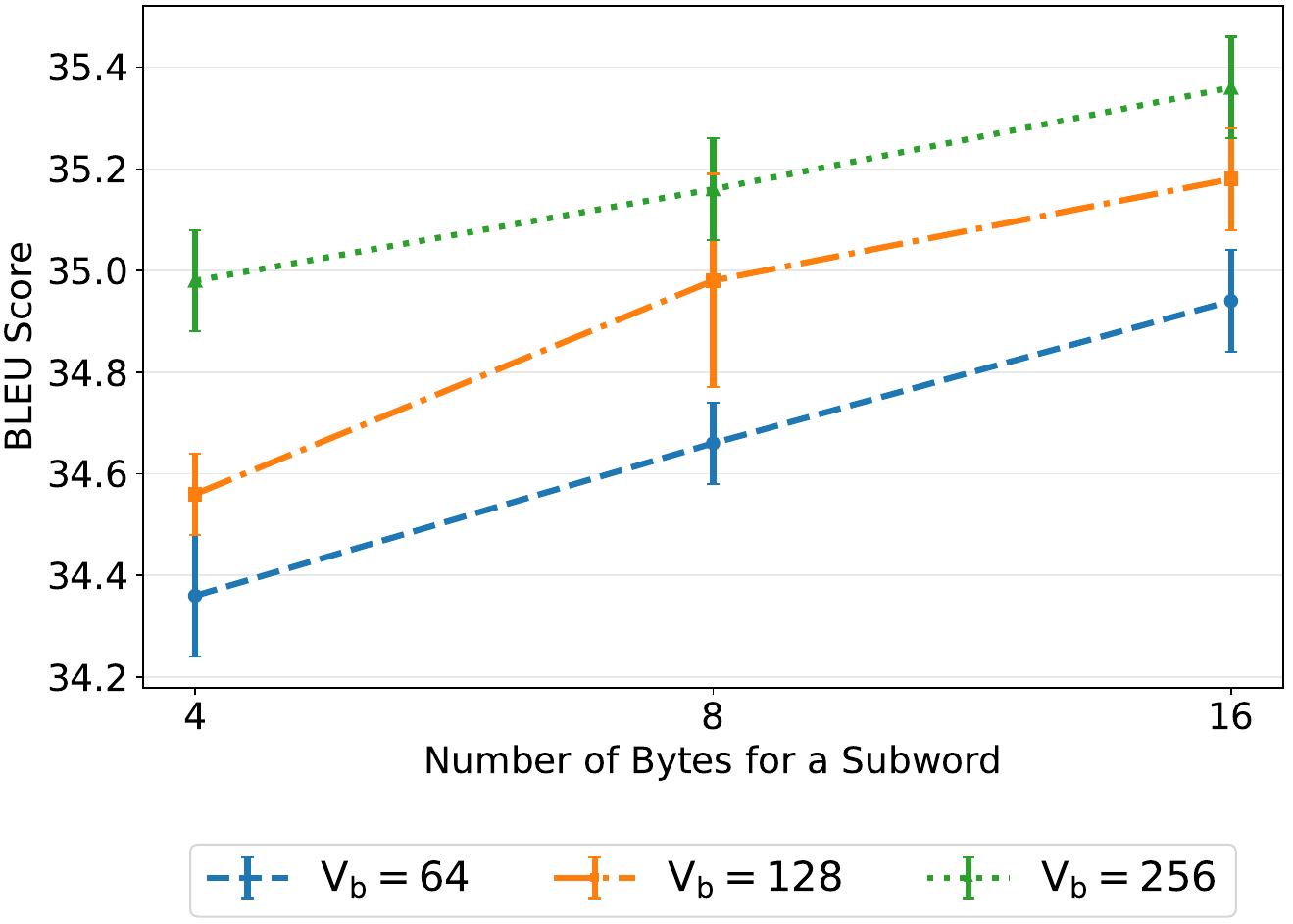}
        \label{fig: byte_dict_subword}}
    \caption{Results on embedding parameters, vocabulary size, and number of bytes per subword. Left: The BLEU scores versus hidden dimension in FFN and embedding parameters. Right: Comparison of mean BLEU scores for different byte vocabulary sizes and different numbers of bytes per subword.}
    \label{fig: BLEU_hidden_dict}
\end{figure*}

To investigate the impact of the byte vocabulary size $V_b$ and number of bytes per subword $n$, we set $V_b$ as ${64, 128, 256}$ and $n$ as ${4, 8, 16}$. Based on the previous experiments, we set the hidden units in the 2-layer FFN to 1024 in {\model}$_{co}$, which provides good performance with a small scale of parameters. We first discuss the model size in terms of embedding parameter numbers in \tableautorefname{~\ref{tab:byte_dict_para}}. All settings have smaller embedding parameter numbers than the original Transformer. We further demonstrate the translation performance under these settings in \figureautorefname{~\ref{fig: BLEU_hidden_dict}} (right). It indicates that increasing $n$ leads to better model performance for a fixed $V_b$. The reason is increasing $n$ results in more possible positions per byte token, providing more information in the aggregated vector. Similarly, when we fix $n$ and increase $V_b$, the increased byte vocabulary diversity makes the aggregated vector more expressive. Therefore, increasing the byte vocabulary size and the number of byte tokens per subword can improve the model's expressiveness, leading to improved performance. Furthermore, \figureautorefname{~\ref{fig: BLEU_hidden_dict}} (right) and \tableautorefname{~\ref{tab:byte_dict_para}} show that models with similar amounts of parameters have similar performance. As long as $V_b$ and $n$ ensure that {\model}$_{co}$ has sufficient expressive ability, the model performance is more related to the number of parameters than to specific $V_b$ and $n$.


\paragraph{Sensitivity analysis on FFN hidden units}
In this experiment, we test the sensitivity of {\model}$_{co}$ on FFN hidden units, because it is one of the major factors for embedding parameters. Here, we set different FFN hidden units as $\{128, 256, 512, 1024, 2048, 4096\}$, with the total embedding parameter numbers of 0.3M, 0.7M, 1.3M, 2.7M, 5.2M, and 10.5M, respectively. The number of embedding parameters and translation BLEU scores are shown on the left of \figureautorefname{~\ref{fig: BLEU_hidden_dict}}. When the numbers of hidden units are 256, 512, and 1024, {\model}$_{co}$ can obtain better performance with fewer parameters. Although the model can still achieve better performance when hidden units are larger than 2048, it does not have advantages over the original transformer on model size.

\subsubsection{Sentiment Analysis}

\paragraph{Dataset and evaluation metrics}
We use IMDb~\cite{maas-EtAl:2011:ACL-HLT2011} and SST2~\cite{socher-etal-2013-recursive} datasets provided by Hugging Face. The detailed preprocessing of the dataset is shown in Appendix~\ref{App:pre}. We use the accuracy for evaluation which is a routine in prior work ~\cite{minaee2019deep, 8249013}. The implementation details are in Appendix~\ref{App: implement}.


\begin{wraptable}{r}{0.4\textwidth}
\vspace{-1\baselineskip}
\small
\centering
\caption{Results on Sentiment analysis.}
\begin{tabular}{lcc} 
\toprule
& IMDb (\%)          & SST2 (\%)\\
\midrule
Subword& 85.6 $\pm$ 0.5 & 81.2 $\pm$ 0.7  \\
$\model_{co}$ & \textbf{85.8 $\pm$ 0.2}  & \textbf{82.5 $\pm$ 0.7}  \\
\bottomrule
\end{tabular}
\label{tab: sentiment}
\end{wraptable}

\paragraph{Main results}




We compare the same BiLSTM models with subword embedding and {\model}$_{co}$. The classification accuracies are shown in \tableautorefname{~\ref{tab: sentiment}}. The results show that {\model}$_{co}$ can replace the conventional subword embedding without hurting the model performance. For SST2, {\model}$_{co}$ even has better performance. The reason for that is the parameters of the conventional subword embedding layer in BiLSTM take a large portion of the model parameters, making the model easily overfitting. In this experiment, {\model}$_{co}$ has smaller embedding parameters, which can address overfitting. We show that {\model} also learns the semantic meaning of subword in~\ref{app:semantic}.


\subsubsection{Language Modeling}

\paragraph{Dataset and evaluation metrics} 
We use the same data as Fairseq did for the language modeling tasks. The dataset we use is WikiText-103. We use the same preprocessing and training settings as the official Fairseq does. The number of samples for training, testing, and validation are 1801350, 3760, and 4358 respectively. We evaluate the language modeling performance with perplexity.



\paragraph{Main results}

For language modeling (LM), our proposed method achieved better performance on perplexity with fewer parameters. The results are shown in \tableautorefname{~\ref{tab:lm}}, which demonstrate that our method \model is an effective and efficient alternative to the traditional subword embedding.



\subsubsection{Federated Learning}
we experiment with federated learning on the SST2 sentiment analysis task using the FedAvg framework\cite{mcmahan2017communication}. In this experiment, we have 20 clients and distribute training samples uniformly to these clients. In training, we sample a part of the clients with the ratio $c$ in every communication round. The results are shown in the following table. We can see that even in the federated learning framework, our \model method is still comparable to subword embedding and can achieve stable results when the number of clients in training varies.

\begin{table*}[t]
    \begin{minipage}[t]{0.41\linewidth}
        \centering
        \small
        \caption[Perplexity of language modeling.]{Perplexity of language modeling for subword embedding and $\mathrm{model}_{co}$.}
        \begin{tabular}{ccc}
        \toprule
                    & \# Parameters & Perplexity \\ \cmidrule(l){1-3} 
        Subword     & 13.7M        & 30.84 \\
        $\mathrm{model}_{co}$ & 10.5M         & 30.55 \\ \bottomrule
        \end{tabular}
        \label{tab:lm}
    \end{minipage}
    \hspace{0.01\linewidth}
    \begin{minipage}[t]{0.56\linewidth}
        \centering
        \small
        \caption{Accuracies for subword embedding and our method for federated learning. $c$ is the participation ratio in each communication round.}
        \begin{tabular}{@{}llllll@{}}
        \toprule
        Embeddings & c=0.2 & c=0.4 & c=0.6 & c=0.8 & c=1.0 \\ \midrule
        Subword & 81.5\% & 80.6\% & 81.1\% & 80.7\% & 80.9\% \\ \midrule
        SEB & 82.0\% & 81.7\% & 81.9\% & 82.4\% & 81.7\% \\ \bottomrule
        \end{tabular}
        \label{tab:fl_results}
    \end{minipage}
\end{table*}

\section{Conclusion}

This paper introduces {\model}, Subword Embedding of Bytes, a novel subword embedding method that defends against privacy leakage attacks based on embedding gradients in federated learning. Different from traditional approaches that learn a large subword embedding matrix, {\model} uses smaller byte embedding matrices or byte one-hot encoding and aggregates byte representations to obtain subword embeddings. With {\model}, attackers cannot retrieve a small set of subwords and generate private text, even with a well-trained large language model. Additionally, we demonstrate that {\model} makes it difficult for attackers to recover private text with embedding gradients in federated learning. Our extensive experiments show that {\model} is effective for machine translation, sentiment analysis, and language modeling tasks without sacrificing model performance or efficiency.

\section{Limitations and Broader impact}

\paragraph{Large language model and more tasks}

Limited to the computation resources, we only experiment with small Transformer models and moderate-size datasets. Moreover, we consider three common tasks such as machine translation, sentiment analysis, and language Modeling to show the effectiveness of our proposed method. The efficiency and effectiveness of our proposed method on large language models as well as other natural language processing tasks still need exploration.

\paragraph{Pretraining exploration}

All models in the experiments are trained from scratch. We have not experimented with the pretraining and finetuing/prompting paradigm with {\model}. However, our proposed {\model} is an effective alternative to subword embedding, and we think that our method is generalizable to popular NLP models, tasks, and training paradigms.
In the future, we will investigate the performance of our proposed method under the pretraining and finetuing/prompting paradigm.

\paragraph{Broader impact}
In this work, we use the IWSLT14~\citep{cettolo2014report}, WMT14~\citep{bojar-etal-2014-findings}, SST2~\citep{socher-etal-2013-recursive}, IMDb~\citep{maas-EtAl:2011:ACL-HLT2011}, and WikiText-103~\citep{merity2016pointer}. All these 5 datasets are widely used pubic datasets in NLP tasks, which do not have personal or sensitive information. For the first four datasets, they are freely used without any license. For the WikiText-103 dataset, it is used under the "Creative Commons Attribution-ShareAlike License". We believe our work does not bring more risk in training and using NLP models. Our intention is to provide practical privacy-preserving collaboration and build trust among clients while maintaining the model's performance and efficiency. We hope this work can motivate more effective defenses, encourage secure and privacy-preserving collaborations in practical applications and have a positive effect on popular large-scale language models.




\bibliography{neurips_2024}
\bibliographystyle{plain}


\appendix
\newpage
\setcounter{section}{0} 
\renewcommand{\thesection}{\Alph{section}}

\section{{\model} Algorithm}
Algorithm \ref{alg:byte_tokenizer} illustrates the detailed steps of our proposed subword to byte sequence mapping in Section~\ref{sec: byte_tokenization}. Algorithm~\ref{alg:aggregation} gives the step-by-step operations to obtain the subword embedding with {\model} based on bytes, given the mapping in Algorithm \ref{alg:byte_tokenizer}. 

\begin{algorithm}[!ht]
    \caption{Construction of Subword to byte sequence mapping}
    \label{alg:byte_tokenizer}
    \DontPrintSemicolon
    \SetAlgoLined
    \newcommand\mycommfont[1]{\ttfamily\textcolor{gray}{#1}}
    \SetCommentSty{mycommfont}
    \SetKwInOut{Input}{Input}
    \SetKwInOut{Output}{Output}
    \Input{Byte vocabulary $\mathcal{V}_b = \{0, 1, \dots, V_b - 1\}$; \\ Subword vocabulary $\mathcal{V}_w = \{w_0, w_1, \dots, w_{V_w - 1}\}$; \\The number of bytes per subword $n$}
    \Output{Mapping: $\mathcal{M}: \mathcal{V}_w \rightarrow (\mathcal{V}_b) ^ {n}$} 
    \For{$w_i \in D_w$}{
        \Repeat{$(b_{i1}, b_{i2}, \dots, b_{in}) \notin \mathcal{M}$}{
            \For{$j \leftarrow 1$ to $n$}{
                Sample $b_{ij} \sim$ \text{Uniform}[0, $V_b$) \tcp*[l]{$\text{P}(b_{ij}) = \frac{1}{V_b}, \quad b_{ij} \in \mathcal{V}_b$}
            }
        }
        Add $w_i \rightarrow (b_{i1}, b_{i2}, \dots, b_{in})$ to $\mathcal{M}$ \\
    }
    \Return{$\mathcal{M}$}
\end{algorithm}

\begin{algorithm}[!ht]
    \caption{Subword embedding with {\model}}
    \label{alg:aggregation}
    \DontPrintSemicolon
    \SetAlgoLined
    \newcommand\mycommfont[1]{\ttfamily\textcolor{gray}{#1}}
    \SetCommentSty{mycommfont}
    \SetKwInOut{Input}{Input}
    \SetKwInOut{Output}{Output}
    \Input{Subword to byte sequence mapping $\mathcal{M}: \mathcal{V}_w \rightarrow (\mathcal{V}_b) ^ {n}$; \\ Subword sequence $S = (w_1, w_2, \dots, w_m)$; \\ A feed-forward network \texttt{FFN}; \\ Embedding matrix is $\mathbf{B} \in \mathbb{R} ^{V_b \times d}$; \\ Subword embedding dimension $d'$}
    \Output{Subword embeddings ${\mathbf{E}'} \in \mathbb{R} ^{m \times d'}$}
    
    \For{$i \leftarrow 1$ to $m$}{
        $(b_{i1}, b_{i2}, \dots, b_{in}) \leftarrow \mathcal{M}[w_i]$\\
    }
    Byte sequence for $S: (b_{11}, \dots, b_{1n}, \dots, b_{m1}, \dots, b_{mn})$ \\

    $\mathbf{E} \in \mathbb{R} ^{mn \times d} \leftarrow$ retrieve $(b_{11}, \dots, b_{1n}, \dots, b_{m1}, \dots, b_{mn})$ from $\mathbf{B}$

    $\Tilde{\mathbf{E}} \in \mathbb{R} ^{m \times nd}$ $\leftarrow$ reshape $\mathbf{E}$ (in a row-major order)


    ${\mathbf{E}'} = \texttt{FFN}(\Tilde{\mathbf{E}}) \in \mathbb{R} ^{m \times d'}$ \\
    \Return{${\mathbf{E}'}$}
\end{algorithm}

\section{ Experimental Details and More Results}

\subsection{Enviormental Settings} 
\label{app:env}
All the programs in our work are implemented using Python 3.9.0, Fairseq~\cite{ott2019fairseq}, PyTorch 1.13.0, and CUDA 11.7. For the hardware environment, we run all codes on a machine with Intel i7-11700K CPU, 64G memory, and NVIDIA GeForce RTX 3080 GPU.

\subsection{Dataset and evaluation metrics}
\label{app:datset}

\paragraph{Translation}

In the translation task, we consider two datasets, one is the medium-size IWSLT14~\citep{cettolo2014report} dataset and a large-scale dataset WMT14~\cite{bojar-etal-2014-findings}. 
We follow the settings as prior work~\citep{shaham2020neural, zhang-xu-2022-byte} and translate German (de) to English (en) in IWSLT14~\citep{cettolo2014report}. The translation of WMT is English (en) to German (de) and the preprocessing is the same as Fairseq~\citep{ott2019fairseq}. 
We use SacreBLEU, case-sensitive, with the 13a tokenizer~\cite{post-2018-call} as the evaluation metric. 

\paragraph{Sentiment Analysis}
We use IMDb~\cite{maas-EtAl:2011:ACL-HLT2011} and SST2~\cite{socher-etal-2013-recursive} datasets provided by Hugging Face. The detailed preprocessing of the dataset is shown in Appendix~\ref{App:pre}. We use the accuracy for evaluation which is a routine in prior work ~\cite{minaee2019deep, 8249013}.

\paragraph{Language modeling}
We use the same data as Fairseq did for the language modeling tasks. The dataset we use is WikiText-103. We use the same preprocessing and training settings as the official Fairseq does. The number of samples for training, testing, and validation are 1801350, 3760, and 4358 respectively. We evaluate the language modeling performance with perplexity.

\subsection{Preprocessing Details}
\label{App:pre}
\paragraph{Translation}
For IWSLT14, there are 166K sentence pairs for training and validation and 5.6K for testing. The vocabulary shared by the source and target languages is built by BPE~\cite{sennrich2015neural} with 10K tokens.

For WMT14, en-de contains 4.5M sentence pairs. Newstest2013 is used for validation and newstest2014 for testing respectively. The merge operation is 32K for BPE and the dictionary is shared by source and target.

\paragraph{Sentiment analysis}
There are 25000 training samples and 25000 testing samples for IMDb. We take 25\% of the training data for validation and the rest for training. For SST2, The training, validation, and test examples in SST2 are 67349, 872, and 1821, respectively. The tokenizer is ``basic\_english'' in the TorchText package. The minimum frequency needed to include a token in the vocabulary is 5. The maximum length of the sentence is 256.

\begin{table*}[ht]
    \centering
    \caption[Cosine similarity of subword embeddings.]{The cosine similarity of the subword embeddings calculated based on \model.}
    \begin{tabular}{ccccccc}
    \toprule
    \textbf{} & \textbf{good} & \textbf{great} & \textbf{funny} & \textbf{bad} & \textbf{worse} & \textbf{boring} \\ 
    \cmidrule(l){2-7} 
    \textbf{good} & 1 & 0.63 & 0.49 & -0.58 & -0.61 & -0.58 \\
    \textbf{great} & 0.63 & 1 & 0.40 & -0.53 & -0.33 & -0.38 \\
    \textbf{funny} & 0.49 & 0.40 & 1 & -0.72 & -0.61 & -0.60 \\
    \textbf{bad} & -0.58 & -0.53 & -0.72 & 1 & 0.72 & 0.85 \\
    \textbf{worse} & -0.61 & -0.33 & -0.61 & 0.72 & 1 & 0.88 \\
    \textbf{boring} & -0.58 & -0.38 & -0.60 & 0.85 & 0.88 & 1 \\ 
    \bottomrule
    \end{tabular}
    \label{tab:cos_sim}
\end{table*}

\subsection{Implementation Details} 
\label{App: implement}
\paragraph{Translation}
The baseline we compare is the transformer with subword embedding~\cite{vaswani2017attention}. Our proposed method only replaces the subword embedding with {\model}. 

For IWSLT14, the encoder and decoder layers are both 6 and have 4 attention heads. The hidden dimension of attention is 512 and the dimension of the feedforward layer is 1024. The optimizer is Adam~\cite{kingma2014adam} with an inverse square root learning rate scheduler, and warm up 4000 steps. The learning rate is $5 \times 10^{-4}$. The total training epochs are 100, and we average the best 5 checkpoints for testing. 

For WMT14, the encoder and decoder layers are both 6 and have 8 attention heads. The hidden dimension of attention is 512 and the dimension of the feedforward layer is 2048. The optimizer is Adam~\cite{kingma2014adam} with an inverse square root learning rate scheduler, and warm up 4000 steps. The learning rate is $5 \times 10^{-4}$. The total training epochs are 100 and we use the early stop if the validation loss does not decrease in 5 epochs. We average the best 5 checkpoints for testing. 

\paragraph{Sentiment Analysis}

 We use 2-layer BiLSTMs for both IMDb and SST2 classification tasks. We keep all model architectures the same for the baseline models and models with our {\model}$_{co}$ except for the embedding parts. The subword embedding dimension is 64 and 256 for IMDb and SST2. The hidden units are 64 and 300 for IMDb and SST2. The hidden dimension of 2-layer FFN in {\model}$_{co}$ is 128 for both datasets. We optimize the model using Adam~\cite{kingma2014adam} and the learning rate is $5 \times 10^{-4}$ for the baseline and our method on both datasets. The best model parameters evaluated on validation data are applied for testing. 

 \paragraph{Language modeling}
 
In this experiment, we also use a two-layer FFN in \model, which has 4096 hidden units. The architecture is $\mathtt{transformer\_lm}$ in the Fairseq framework. We share the input and output embedding in the encoder and the other hyperparameters and settings are the same as Fairseq.

\subsection{Analysis for Semantic Meaning}
\label{app:semantic}

In this section, we will analyze whether the derived subword embeddings from our method can truly encode the meaning of the words in the embedding space. To experiment on this aspect, we mainly calculate the cosine similarity between two word embeddings obtained based on our method \model. We list some examples in IMDb sentiment analysis in \tableautorefname{~\ref{tab:cos_sim}}.

The cosine similarity demonstrates that the subword embedding of our proposed \model will learn the semantic meaning from the task. For example, positive words (good, great, and funny) have positive and high-value similarities with each other, which is also the same case for all negative words (bad, worse, and boring). However, all the negative-positive pairs have negative similarities, which means the subword embedding of our proposed \model can automatically learn the semantic meaning.

\subsection{Analysis for Space Complexity}
\label{app:space}
We analyze the space complexity in the experiments. We mainly take the translation on IWSLT14 and sentiment analysis as examples. Transformer model for translation on IWSLT14 de-en with 256 hidden units in our method \model as the translation results are close to the traditional subword embedding.

The tables below present the sizes of both the entire model's parameters and the embedding layer for translation on IWSLT and sentiment analysis, respectively. The numbers in ``()'' represent the percentage reduction achieved by our method compared to the subword model.

\begin{table*}[htbp]
\small
    \centering
    \caption{Transformer model for translation on IWSLT14 de-en.}
    \begin{tabular}{cccc}
    \toprule
    \text{\# Params} & \text{Whole model} & \text{Embedding} & \text{BLEU} \\ 
    \midrule
    Subword & 37M & 5.2M & 34.54 $\pm$ 0.10 \\
    {\model}$_{co}$ & 33M ($\downarrow$ 12\%) & 0.7M ($\downarrow$ 94\%) & 34.62 $\pm$ 0.12 \\ \bottomrule
    \end{tabular}
    
    \label{tab:tran_space}
\end{table*}

\begin{table*}[htbp]
\small
    \centering
    \caption{Sentiment analysis on SST2}
    \begin{tabular}{cccc}
    \toprule
    \text{\# Params} & \text{Whole model} & \text{Embedding} & \text{Accuracy} \\ 
    \midrule
    Subword & 5.9M & 2.4M & 81.2 $\pm$ 0.7 \\
    {\model}$_{co}$ & 3.8M ($\downarrow$ 36\%) & 0.8M ($\downarrow$ 68\%) & 82.5 $\pm$ 0.7 \\ \bottomrule
    \end{tabular}
    
    \label{tab:sst2_space}
\end{table*}

\begin{table*}[htbp]
\small
    \centering
    \caption{Sentiment analysis on IMDb}
    \begin{tabular}{cccc}
    \toprule
    \text{\# Params} & \text{Whole model} & \text{Embedding} & \text{BLEU} \\ 
    \midrule
    Subword & 1.5M & 1.5M & 85.6 $\pm$ 0.5 \\
    {\model}$_{co}$ & 0.4M ($\downarrow$ 72\%) & 0.5M ($\downarrow$ 80\%) & 85.8 $\pm$ 0.2 \\ \bottomrule
    \end{tabular}
    
    \label{tab:imdb_space}
\end{table*}

In all of these tasks, our method $\model_{co}$ can decrease the space complexity, which shows the ability of our method to reduce the model size. In scenarios where model training is necessitated on a device with limited memory, it will be better to make the model smaller while keeping the model's performance.

\subsection{Comparison with Gradient Prune Defense}
\label{app:gradient_defense}

We compare the defense method of gradient pruning in the FILM attack for batch size = $8, 16, 32$. Tables~\ref{tab:gradient_prune_8},\ref{tab:gradient_prune_16}, and \ref{tab:gradient_prune_32} show the precision and recall for gradient pruning and our method. Even without pruning, our method has a very low recall compared to FILM on subword embeddings when all batch sizes we experimented on, which shows the effectiveness of our defense.

\begin{table*}[ht]
\small
\centering
\caption{Defense results on precision and recall for batch size is 8.}
\begin{tabular}{@{}lllll@{}}
\toprule
\multicolumn{1}{c}{\multirow{2}{*}{Prune ratio}} & \multicolumn{2}{c}{Precision} & \multicolumn{2}{c}{Recall} \\ \cmidrule(l){2-5} 
\multicolumn{1}{c}{} & Subword & SEB & Subword & SEB \\ \midrule
0 & 1 & 1 & 1 & 0.003 \\
0.9 & 1 & 1 & 1 & 0.003 \\
0.99 & 1 & 1 & 1 & 0.003 \\
0.999 & 1 & 1 & 0.53 & 0.003 \\
0.9999 & 1 & 0.46 & 0.08 & 0.003 \\ \bottomrule
\end{tabular}
\label{tab:gradient_prune_8}
\end{table*}

\begin{table*}[ht]
\small
\centering
\caption{Defense results on precision and recall for batch size is 16.}
\begin{tabular}{@{}lllll@{}}
\toprule
\multicolumn{1}{c}{\multirow{2}{*}{Prune ratio}} & \multicolumn{2}{c}{Precision} & \multicolumn{2}{c}{Recall} \\ \cmidrule(l){2-5} 
\multicolumn{1}{c}{} & Subword & SEB & Subword & SEB \\ \midrule
0 & 1 & 1 & 1 & 0.005 \\
0.9 & 1 & 1 & 1 & 0.005 \\
0.99 & 1 & 1 & 1 & 0.005 \\
0.999 & 1 & 1 & 0.49 & 0.005 \\
0.9999 & 1 & 0.50 & 0.06 & 0.005 \\ \bottomrule
\end{tabular}
\label{tab:gradient_prune_16}
\end{table*}

\begin{table*}[!ht]
\small
\centering
\caption{Defense results on precision and recall for batch size is 32.}
\begin{tabular}{@{}lllll@{}}
\toprule
\multicolumn{1}{c}{\multirow{2}{*}{Prune ratio}} & \multicolumn{2}{c}{Precision} & \multicolumn{2}{c}{Recall} \\ \cmidrule(l){2-5} 
\multicolumn{1}{c}{} & Subword & SEB & Subword & SEB \\ \midrule
0 & 1 & 1 & 1 & 0.009 \\
0.9 & 1 & 1 & 1 & 0.009 \\
0.99 & 1 & 1 & 1 & 0.009 \\
0.999 & 1 & 0.99 & 0.51 & 0.009 \\
0.9999 & 1 & 0.47 & 0.07 & 0.009 \\ \bottomrule
\end{tabular}
\label{tab:gradient_prune_32}
\end{table*}

\section{Discussion of Frequency Analysis}
\label{app:freq_analysis}

Frequency analysis is useful in cryptanalysis. For simple substitution ciphers, there is a characteristic distribution of letters that is roughly the same for almost all samples of that language. Frequency analysis uses the characteristic distributions of the plaintext and ciphertext to guess the mapping between them. 


In the scenario of the threat model and our proposed method, the attacker only knows the gradients, model parameters and the mapping from subword to byte sequence. Based on these, the attacker can only get the information about distinct byte candidates which are updated in training. The attacker cannot determine the frequencies of the bytes based on the gradients. 

To demonstrate the effectiveness of our method, we further assume the attacker have the information about the frequency of each byte. The goal of the attacker is to get the plaintext, given the plaintext to ciphertext mapping, and the characteristic distributions of the ciphertext.

To infer the plaintext, the attacker needs to know the order for combining all of the byte candidates and then use the ciphertext to plaintext mapping to obtain the original text. However, the attacker only knows a bag of bytes without ordering. It is difficult to infer the correct combination of the bytes as possible combinations of bytes is extremely large.

\end{document}